\documentclass{INTERSPEECH2023}

\interspeechcameraready

\usepackage{todonotes}

\usepackage{tabularx}
\usepackage{multicol,multirow}
\usepackage{color, colortbl}
\definecolor{Gray}{gray}{0.9}

\title{Semantic enrichment towards efficient speech representations} 
\name{
    G. Laperrière$^{1,2}$, 
    H. Nguyen$^1$, 
    S. Ghannay$^2$, 
    B. Jabaian$^1$, 
    Y. Estève$^1$}
\address{
    $^1$LIA - Avignon Université, France \\
    $^2$LISN - CNRS/Université Paris-Saclay, France}
\email{
    $^1$firstname.lastname@univ-avignon.fr, 
    $^2$firstname.lastname@lisn.upsaclay.fr}

\begin{document}

\maketitle

\begin{abstract}
\vspace{-0.05cm}
Over the past few years, self-supervised learned speech representations have emerged as fruitful replacements for conventional surface representations when solving Spoken Language Understanding (SLU) tasks.
Simultaneously, multilingual models trained on massive textual data were introduced to encode language agnostic semantics.
Recently, the SAMU-XLSR approach introduced a way to make profit from such textual models to enrich multilingual speech representations with language agnostic semantics.
By aiming for better semantic extraction on a challenging Spoken Language Understanding task and in consideration with computation costs, this study investigates a specific in-domain semantic enrichment of the SAMU-XLSR model by specializing it on a small amount of transcribed data from the downstream task.
In addition, we show the benefits of the use of same-domain French and Italian benchmarks for low-resource language portability and explore cross-domain capacities of the enriched SAMU-XLSR. 
\end{abstract}

\vspace{-0.15cm}
\section{Introduction}
\label{sec:intro}
\vspace{-0.05cm}




Spoken language understanding (SLU) consists of various Natural Language Processing tasks that extract semantics from speech~\cite{tur2011spoken}, such as call routing, named entity recognition from speech, or slot filling tasks in the context of human-machine dialogue. 
This work focuses on end-to-end neural approaches for speech-to-concept, one of the most challenging SLU tasks.
It distinguishes itself from conventional cascade approaches~\cite{liu2020mockingjay,liu2021tera} by using a single neural model to directly extract the semantics from speech signals~\cite{ghannay2018end, haghani2018audio,serdyuk2018towards} with the advantages of: (1) jointly optimizing the Automatic Speech Recognition (ASR) and Natural Language Understanding (NLU) parts, and (2) mitigating error propagation. 
Nonetheless, end-to-end models' main challenge resides in the lack of bimodal annotated data, i.e. audio speech recordings with semantic manual annotations. 
To overcome it, transfer learning techniques~\cite{bhosale2019end,Caubriere2019,huang2020leveraging} and artificial augmentation of training data with speech synthesis~\cite{desot2020corpus,lugosch2020using} have been proposed.


In this paper, we investigated a recently proposed approach to remedy the aforementioned problem with the use of self-supervised learning (SSL) for the SLU task. 
SSL models, which are pre-trained from huge amounts of unlabelled data, have lately become very trendy as they show promising results in a wide range of speech tasks~\cite{baevski2020wav2vec,devlin-etal-2019-bert} when substantially alleviating the need of costly annotated speech data. 
At the same time, similar ideas had been successfully applied to text to allow semantics information extraction~\cite{devlin-etal-2019-bert,feng2022language}. 
Several attempts to unify both textual and speech modalities can be found in~\cite{huang2020leveraging,agrawal2022tie,muller2021pursuit}. 
Inspired by this new challenge, ~\cite{khurana2022samu} proposed a framework named SAMU-XLSR (Semantically-Aligned Multimodal Utterance-level Cross-Lingual Speech Representation) which produces a semantically-aligned multimodal and multilingual sentence-level representation. 
To do so, the authors combined the well-known multilingual frame-level speech representation learning model XLS-R~\cite{xlsr} with the Language Agnostic BERT Sentence Embedding generator LaBSE~\cite{feng2022language}. 
More interestingly, ~\cite{laperriere2022use} shows that SAMU-XLSR can also be used as a frame-level speech encoder for a challenging end-to-end SLU task when they find that this model might create semantically aware frame-level speech representations.   


This study shows that specializing SAMU-XLSR representations, by exploiting a small amount of transcribed data without costly semantic annotation, offers very strong semantics extraction enhancements. 
Another main discovery lies in the scoring equivalence obtained with computational cost-effective experiments following layer-wise analysis of enriched SSL models. 
We finally investigate how different portabilities on same-domain or same-language data could be beneficial in order to make the semantically enriched representations more accurate.

\vspace{-0.1cm}
\section{SAMU-XLSR}
\label{sec:samu}
\vspace{-0.05cm}

Self-supervision for speech representations has emerged recently as a strong alternative to conventional handcrafted approaches such as filter-bank and MFCC features, by being capable of leveraging large amounts of unlabelled speech data to learn useful latent speech features.
Outstanding representatives like wav2vec 2.0~\cite{baevski2020wav2vec}, HuBERT~\cite{hsu2021hubert}, and WavLM~\cite{chen2022wavlm} have proven their effectiveness in various speech tasks such as ASR~\cite{baevski2020wav2vec}, speaker verification~\cite{chen2022large,chen22g_interspeech} and emotion recognition~\cite{macary2021use,pepino21_interspeech}.
While being designed to learn speech embeddings at acoustic frame-level, i.e., for short speech segments of 20 milliseconds, \cite{khurana2022samu} proposed the SAMU-XLSR framework to modify the speech embeddings of a pre-trained SSL model in order to capture semantic information, by exploiting audio/text paired data independent of the downstream tasks.

Higher-level semantics were proven useful for cross-lingual speech-to-text mining and zero-shot speech-to-text translation settings, hinting that other speech tasks such as SLU might benefit from this kind of speech representations.

\begin{figure}[htb!]
\centering
    \includegraphics[width=0.97\columnwidth]{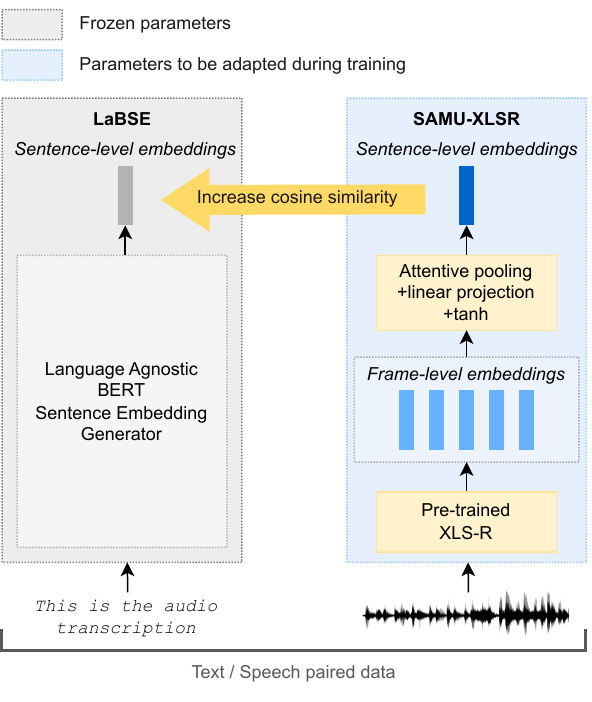} 
    \vspace{-0.35cm}
    \caption{Training process of SAMU-XLSR}
    \label{fig:samu}
    \vspace{-0.6cm}
\end{figure}

In detail, as shown in figure~\ref{fig:samu}, SAMU-XLSR learns to generate a sentence-level embedding using an Attentive Pooling block taking as input the frame-level contextual representations of a pre-trained XLS-R model~\cite{xlsr}.
On the same semantic space, this speech sentence embedding is pulled towards a textual sentence embedding generated by the pre-trained LaBSE model~\cite{feng2022language}, by a cosine similarity loss.
This way, the weights of the speech components (the pre-trained XLS-R and the Attentive Pooling block) are updated to predict the text embeddings provided by the frozen LaBSE model for the corresponding transcripts.
Furthermore, since both XLS-R and LaBSE are multilingual models, SAMU-XLSR is inherently multimodal and cross-lingual.
After the training process, the SAMU-XLSR model can produce either frame-level or sentence-level embeddings for the downstream tasks.

\vspace{-0.1cm}
\section{MEDIA and Italian PortMEDIA datasets}

In this study, we used two datasets of recorded phone calls for hotel booking, in two languages: Italian (part of the PortMEDIA corpus \cite{lefevre-etal-2012-leveraging}) and French (MEDIA corpus \cite{BonneauMaynard2005}). 
They are issued from the \textsl{MEDIA Evaluation Package}\footnote{\url{http://catalog.elra.info/en-us/repository/browse/ELRA-E0024/}} \footnote{International Standard Language Resource Number: 699-856-029-354-6} distributed by ELRA and freely accessible for academic research. All recorded participants gave their permission to create and distribute the data collections.
These datasets are dedicated to semantic information extraction from speech in the context of human-machine dialogues collected by using the Wizard-of-Oz method. 
Only the users' turns are annotated with both manual transcriptions and complex semantic annotations, and used in this study.
Table~\ref{tab:stat} presents the description of both corpora in terms of hours and word occurrences.

\begin{table}[!ht]
    \begin{center}
        \caption{Data distribution of MEDIA and Italian PortMEDIA.}
        \vspace{-0.2cm}
        \resizebox{\columnwidth}{!}{
        \begin{tabular}{ c  c | c | c | c |}
            \cline{3-5}    
            \multicolumn{2}{c |}{} & \textbf{Train} & \textbf{Dev} & \textbf{Test} \\
            \hline   
            \multirow{2}{*}{\textbf{Hours}} & MEDIA & 10h52m & 01h13m & 03h01m \\
            & Italian PortMEDIA & 07h18m & 02h32m & 04h51m \\
            \hline   
            \multirow{2}{*}{\textbf{Words}} & MEDIA & 94.5k & 10.8k & 26.6k \\
            & Italian PortMEDIA & 21.7k & 7.7k & 14.7k \\
            \hline  
        \end{tabular}
        \label{tab:stat}
        }
    \vspace{-0.6cm}
    \end{center}
\end{table}

The Italian PortMEDIA corpus~\cite{lefevre-etal-2012-leveraging} \footnote{\url{http://www.elra.info/en/projects/archived-projects/port-media/}} is made of 604 dialogues from more than 150 Italian speakers.
Its semantic dictionary includes 83 basic attributes and 19 specifiers~\cite{bonneau-maynard-etal-2006-results}.
This corpus is only available with ``full" semantic tags, which include specifiers and modifiers, totaling 139 different concepts.

The French MEDIA corpus, containing 1258 dialogues from around 250 speakers, was used for more in-domain data during SAMU-XLSR specializations, and to conduct experiments on SLU language portability from French to Italian.

The end-to-end SLU models we built in this work predict a transcription enriched with semantic labels, such as: \textsl{I \textnormal{{\textless}reservation\textgreater} would like to book \textnormal{\textgreater} \textnormal{{\textless}room-number\textgreater} one \textnormal{\textgreater} \textnormal{{\textless}room-type\textgreater} double room \textnormal{\textgreater}}. 
Two metrics are commonly used for this benchmark: the Concept Error Rate (CER) and the Concept-Value Error Rate (CVER). 
Both are computed similarly to the Word Error Rate (WER) except that CER only takes into account the concepts occurrences, while CVER considers the correctness of the complete concept/value pair (\textsl{\textnormal{{\textless}room-number\textgreater} one \textnormal{\textgreater}}). 
We also evaluate the transcription (excluding concepts) generated by our models with the WER. 

\vspace{-0.2cm}
\section{General SLU architecture}
\label{sec:generalarch}
\vspace{-0.05cm}

\begin{figure}[htb!]
\centering
    \vspace{-0.3cm}
    \includegraphics[width=0.41\columnwidth]{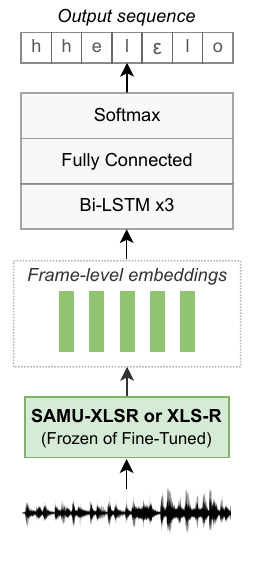}
    \vspace{-0.6cm}
    \caption{SLU end-to-end architecture.} 
    \label{fig:slu}
    \vspace{-0.3cm}
\end{figure}

The end-to-end model used for our SLU task (Figure~\ref{fig:slu}) consists of a frozen or fine-tuned speech encoder (XLS-R or one of the SAMU-XLSRs presented in section \ref{sec:specializing}), followed by 3 bi-LSTM layers of 1024 neurons for segments contextualization. 
Then follows a fully connected layer of the same dimension activated with LeakyReLU and a softmax function.  

We optimize the CTC greedy loss function by using an Adam optimizer with $learning\_rate=0.0001$ for the speech encoders and Bi-LSTM layers, and an Adadelta optimizer with $learning\_rate=1.0$ for the fully connected layer.

We also tested other RNNs (Bi-GRU, Li-GRU, LSTM) with different numbers of blocks ($0$ to $4$) and $0$ to $4$ fully connected layers, all for multiple neural dimensions ($512$, $2048$).

The models presented in this paper learn $387.8$M parameters in 16.5 hours on Italian PortMEDIA and 27 hours on MEDIA with a single v100-32G GPU, when fully fine-tuned for $100$ epochs.
Freezing speech encoders reduces the parameters by $316.3$M and running time by around 6.5 hours on Italian PortMEDIA (-$40$\%) and 12 hours (-$44$\%) on MEDIA. 

Note that all following SLU experiments need to take in consideration a $0.4$\% possible variation of Error Rates, estimated from 5 training runs on PortMEDIA with different seeds. 

\vspace{-0.2cm}
\section{Contributions}
\label{sec:contrib}
\vspace{-0.05cm}

In this paper, we propose a semantically enriching specialization of the SAMU-XLSR representations to an in-domain downstream task for close languages with very few hours of speech.
Our investigation also focuses on a computational cost-effective way to approach a challenging SLU task.
We then analyze its new semantics extraction abilities considering language portability, by training our SLU model for 100 epochs on the French MEDIA dataset, followed by 100 epochs on the Italian PortMEDIA dataset.
These experiments will be named FR$\rightarrow$IT.

The main limitation of our study lies in the similarity of French and Italian languages, MEDIA and Italian PortMEDIA being the currently only existing multi-lingual datasets for same-semantics extraction. 

\vspace{-0.1cm}
\subsection{Specializing SAMU-XLSR} 
\label{sec:specializing}
\vspace{-0.05cm}

We explored a ``specialization" of the SAMU-XLSR model. To do so, we fine-tuned the model presented in Figure \ref{fig:samu} on our downstream task's data without semantic tags, to make SAMU-XLSR speech representations closer to their corresponding LaBSE text representations, as done in the original work~\cite{khurana2022samu}. This way, we expect to adapt the semantic alignment of SAMU-XLSR representations to our SLU task, thanks to in-domain transcriptions. These specialized speech encoders are then used in our end-to-end SLU model as shown by Figure \ref{fig:slu}.



We specialized the SAMU-XLSR model on either MEDIA, Italian PortMEDIA, or the combination of the two, for $100$ epochs with Adam optimizer's initial learning rate set to $0.0001$.
These three models are respectively named: SAMU-XLSR $_{FR}$, SAMU-XLSR $_{IT}$, and SAMU-XLSR $_{IT \oplus FR}$, while SAMU-XLSRs refers to the original non-specialized SAMU-XLSR and all the models derived from it. 
During validation, Italian PortMEDIA's dev set was used on SAMU-XLSR $_{IT}$ and SAMU-XLSR $_{IT \oplus FR}$, whereas MEDIA's dev set was used on SAMU-XLSR $_{FR}$.
All three models have $316.2$M parameters, and were trained for respectively $20$ hours, $13.3$ hours, and $33.3$ hours in total, on a single v100-32G GPU.

In Table \ref{tab:base_res_frozen}, we studied the impact of these specializations on the Italian PortMEDIA SLU task. 
When freezing the different speech encoders, specialized SAMU-XLSRs significantly outperform both XLS-R and SAMU-XLSR.
We insist on the fact that freezing the speech encoder results in a model of only $71.5$M parameters to be trained in comparison with the $387.8$M parameters to be updated while fine-tuning it.

\begin{table}[!ht]
    \begin{center}
    \vspace{-0.16cm}
    \caption{Results (\%) for Italian PortMEDIA while freezing speech encoders.}
    \vspace{-0.5cm}
    \resizebox{0.74\columnwidth}{!}{
        \begin{tabular}{ l | c | c | c |}
            \cline{2-4}   
            & \textbf{WER} & \textbf{CER} & \textbf{CVER} \\
            \hline
            XLS-R & $33.7$ & $42.1$ & $57.1$ \\
            SAMU-XLSR & $31.5$ & $33.6$ & $49.0$ \\
            SAMU-XLSR $_{IT}$ & $19.6$ & $30.3$ & $42.9$ \\
            SAMU-XLSR $_{IT \oplus FR}$ & \textbf{18.7} & \textbf{29.4} & \textbf{41.6} \\
            \hline  
        \end{tabular}
        \label{tab:base_res_frozen}
        }
    \vspace{-0.6cm}
    \end{center}
\end{table}

Table~\ref{tab:base_res_finetuned} demonstrates no further improvements when fine-tuning SAMU-XLSR $_{IT}$ during the SLU training process. 
This can be explained because the speech encoder has been trained on the downstream training data to extract the general semantics of a sentence in the specialization process.
However, considering the very small amount of Italian PortMEDIA data, fine-tuning SAMU-XLSR $_{IT \oplus FR}$, which was specialized on more in-domain cross-lingual data, resulted in significant score improvements.  
Our end-to-end SLU models also surpass our previous work~\cite{laperrierelrec}, scoring $18.5\%$ CER on the MEDIA  ``full" mode with a multilingual wav2vec 2.0, instead of $26.1$\% with the French LeBenchmark wav2vec 2.0, and $20.3$\% with additional CommonVoice French data.

\begin{table}[!ht]
    \begin{center}
    \vspace{-0.15cm}
    \caption{Results (\%) for Italian PortMEDIA while Fine-Tuning speech encoders.}
    \vspace{-0.48cm}
    \resizebox{0.74\columnwidth}{!}{
        \begin{tabular}{ l | c | c | c |}
            \cline{2-4}  
            & \textbf{WER} & \textbf{CER} & \textbf{CVER} \\
            \hline
            XLS-R & $18.1$ & $29.6$ & $41.5$ \\
            SAMU-XLSR & $15.4$ & $26.6$ & $39.2$ \\
            SAMU-XLSR $_{IT}$ & $15.7$ & $26.8$ & $39.5$ \\
            SAMU-XLSR $_{IT \oplus FR}$ & \textbf{14.5} & \textbf{25.6} & \textbf{37.6} \\
            \hline  
        \end{tabular}
        \label{tab:base_res_finetuned}
        }
    \vspace{-0.8cm}
    \end{center}
\end{table}

\vspace{-0.1cm}
\subsection{Layer-wise speech encoders analysis}
\label{sec:evolution}
\vspace{-0.05cm}

\begin{figure}[htb!]
\centering
    \vspace{-0.1cm}
    \includegraphics[width=0.99\columnwidth]{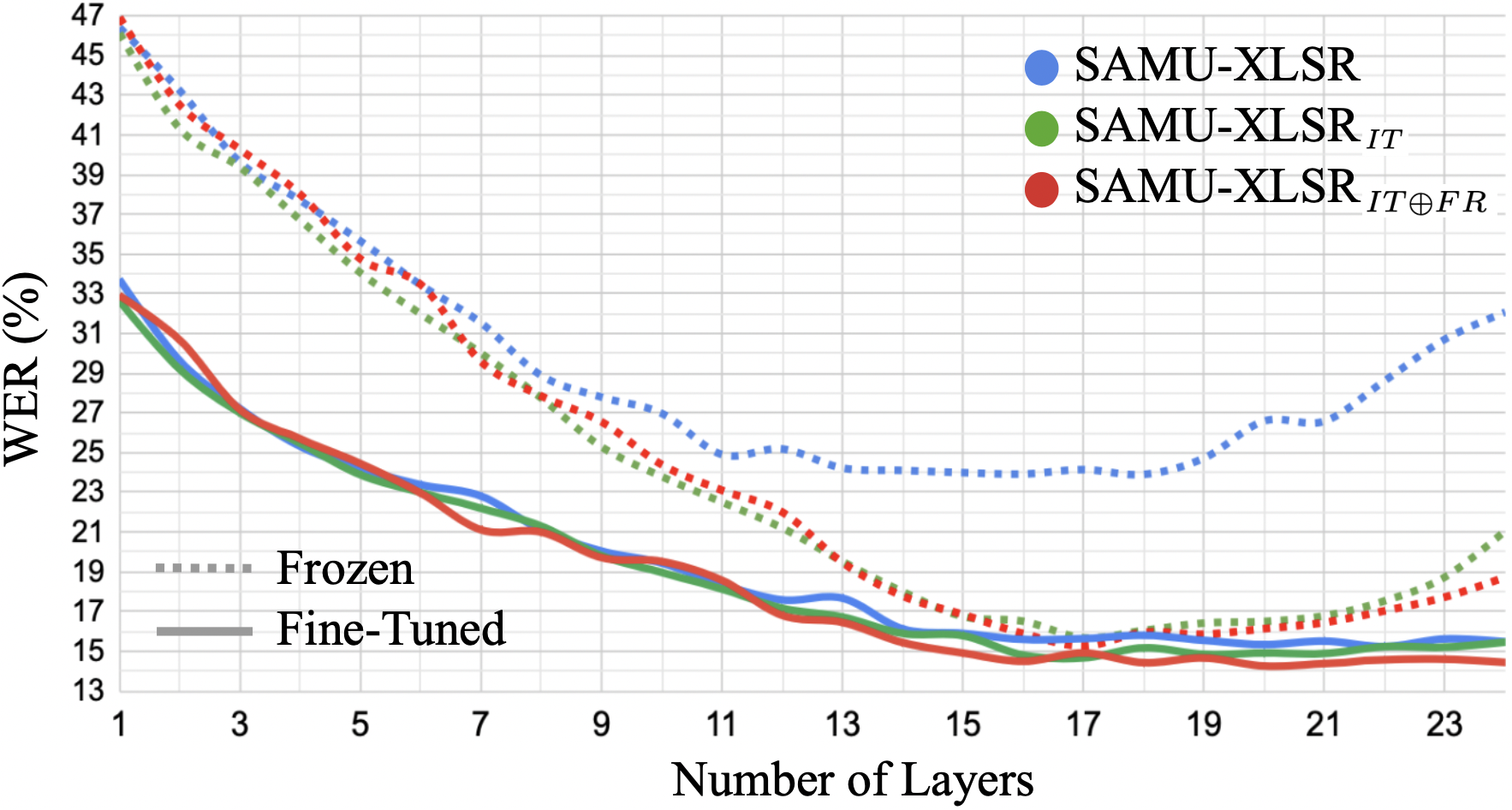}
    \vspace{-0.3cm}
    \caption{Layer-wise WER analysis of specialized SAMU-XLSRs during the Italian PortMEDIA SLU task.} 
    \label{fig:wer}
    \vspace{-0.5cm}
\end{figure}

\begin{figure}[htb!]
\centering
    \includegraphics[width=0.99\columnwidth]{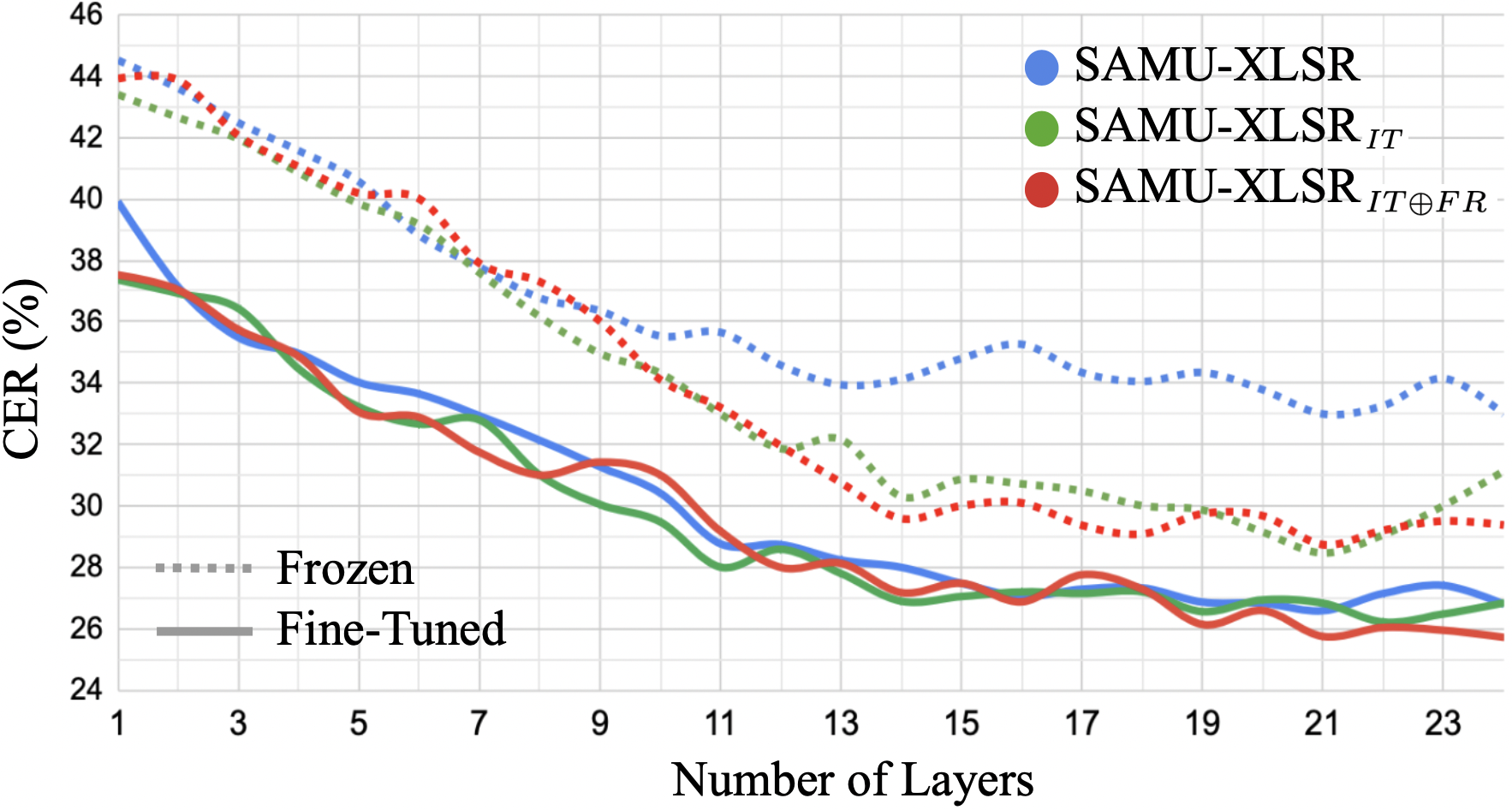}
    \vspace{-0.3cm}
    \caption{Layer-wise CER analysis of specialized SAMU-XLSRs during the Italian PortMEDIA SLU task.} 
    \label{fig:cer}
    \vspace{-0.2cm}
\end{figure}

Figures \ref{fig:wer} and \ref{fig:cer} present respectively a WER and CER layer-wise analysis of specialized SAMU-XLSRs, compared to the original SAMU-XLSR, by removing the upper layers of each encoder, one by one, to extract our speech embeddings.
The kept layers of the encoders are frozen for the ``Frozen" architecture, or fine-tuned by supervision to solve the Italian PortMEDIA SLU task, leading to the ``Fine-Tuned" results. 

Fine-Tuning specialized speech encoders during Italian PortMEDIA training leads to no significant improvement.
However, specialized SAMU-XLSRs generate far better representations than SAMU-XLSR in terms of semantic (CER) and linguistic (WER) encoding, when Frozen for the SLU task. 
As we already know from~\cite{laperriere2022use}, SAMU-XLSR captures and encodes semantics from the speech until its top layers thanks to LaBSE semantically-aligned embeddings. 
Yet, LaBSE being Language Agnostic, linguistic information is less well represented in the last SAMU-XLSRs layers, even when specialized. 

The most important observation appears in Figure \ref{fig:wer}: freezing a specialized SAMU-XLSR with only 17 layers leads to a WER of $15.3$\%, being identical to the $15.4$\% of WER given by fine-tuning a full 24 layers original SAMU-XLSR. 
This result was obtained thanks to the specializations made with in-domain data transcriptions, neither aiming for an ASR nor our final SLU task but only aiming to inject semantics into the speech representations with the use of LaBSE. 

\vspace{-0.1cm}
\subsection{Portability}
\vspace{-0.1cm}

This section explores the impact of SAMU-XLSR's specialization on in-domain cross-lingual experiments (FR$\to$IT), followed by cross-domain same-language experiments with the Italian CommonVoice and French PortMEDIA datasets.

\vspace{-0.2cm}
\subsubsection{Cross-Lingual}
\label{sec:porta}
\vspace{-0.07cm}


New state-of-the-art Italian PortMEDIA CER results were obtained with language portability experiments (Table \ref{tab:portab_res}), by fine-tuning the speech encoders on both MEDIA then Italian PortMEDIA \textbf{FR$\to$IT}.
SAMU-XLSR $_{IT \oplus FR}$ led to $25.1$\% of CER, instead of $26.2$\% obtained with a bigger architecture and the non-specialized SAMU-XLSR~\cite{laperriere2022use}.
With portability experiments, SAMU-XLSR $_{IT \oplus FR}$ significantly outperforms SAMU-XLSR $_{FR}$ in terms of WER and CVER thanks to its specialization on the combined in-domain French and Italian transcriptions. 
The $100$ pre-training epochs done on French data allows the model to learn more about the task semantics but degrades Italian transcriptions, going from at best $14.5$\% of WER with Italian-only training to $16.4$\% of WER with FR$\to$IT fully fine-tuned training.

Freezing speech encoders resulted in far better CER for FR$\to$IT trainings with SAMU-XLSR $_{IT \oplus FR}$ ($26.5$\% CER if not considering SAMU-XLSR $_{IT \oplus FR}$ $_{(17)}$) than for Italian-only trainings with SAMU-XLSR $_{IT \oplus FR}$ ($29.4$\% CER).

\begin{table}[ht]
    \begin{center}
        \vspace{-0.25cm}
        \caption{Results (\%) for Language portability (FR$\rightarrow$IT).}
        \vspace{-0.2cm}
        \resizebox{\columnwidth}{!}{
        \begin{tabular}{ l  l | c | c | c |}
            \cline{3-5}   
            & & \textbf{WER} & \textbf{CER} & \textbf{CVER} \\
            \hline
            \multirow{5}{*}{Frozen} 
            & XLS-R & $26.4$ & $34.9$ & $49.0$ \\
            & SAMU-XLSR & $32.0$ & $30.8$ & $48.7$ \\
            & SAMU-XLSR $_{FR}$ & $31.4$ & $31.0$ & $49.4$ \\
            & SAMU-XLSR $_{IT \oplus FR}$ & \textbf{17.3} & \textbf{26.5} & \textbf{39.2} \\
            \rowcolor{Gray}
            & SAMU-XLSR $_{IT \oplus FR}$ $_{(17)}$ & \textbf{15.9} & \textbf{25.6} & \textbf{38.0} \\
            \hline
            \multirow{4}{*}{Fine-Tuned} 
            & XLS-R & $20.7$ & $27.3$ & $41.1$ \\
            & SAMU-XLSR & $17.5$ & $25.5$ & $38.4$ \\
            & SAMU-XLSR $_{FR}$ & $17.8$ & 25.2 & $39.1$ \\
            & SAMU-XLSR $_{IT \oplus FR}$ & \textbf{16.4} & \textbf{25.1} & \textbf{38.1} \\
            \hline  
        \end{tabular}
        }
    \label{tab:portab_res}
    \vspace{-0.63cm}
    \end{center}
\end{table}

In Figure~\ref{fig:wer}, we noticed that a frozen SAMU-XLSR $_{IT \oplus FR}$ with only \textbf{17 layers} scored almost the best obtained WER. 
Therefore, we wanted to explore the effectiveness of this far smaller model to the FR$\to$IT task, naming the model SAMU-XLSR $_{IT \oplus FR}$ $_{(17)}$ in Table~\ref{tab:portab_res}.
Training for only 9.5 hours on a single v100-32G GPU impressively led to the best WER and CVER scores for portability experiments while approaching the best 25.1\% CER obtained with fully fine-tuned speech encoders in $39$ hours, and equalizing the $25.6$\% CER of Italian-only experiments obtained in 11 hours.

\vspace{-0.2cm}
\subsubsection{Cross-Domain}
\vspace{-0.07cm}

This section aimed to determine whether the specializations degraded or improved significantly the scorings, thus not to give relevant results for both tasks. 
This is why only inferences were made, instead of full trainings. 

Considering completely out-of-domain tasks, we chose to test our Italian PortMEDIA models with \textbf{CommonVoice Italian} dataset~\footnote{https://commonvoice.mozilla.org/fr/datasets}, a corpus of recorded broadcast news unrelated to hotel booking.
Being devoid of semantic annotations, we only evaluated its ASR performances with Character and Word Error Rates (ChER and WER). 
Experiments resulted in almost $100$\% WER for all speech encoders, meaning that our SLU models could not transcribe words from far cross-domain data. 
Even ChER, expected lower due to same-phonemic speech, scored $90.7$\% by fine-tuning XLS-R, against at most $98$\% with specialized SAMU-XLSRs which lost their initial abilities to transcribe broadcast Italian, being specialized on phone calls audio.

To pursue this analysis on close-domain tasks despite the lack of fitting Italian data, we referred to the \textbf{French PortMEDIA} dataset, jointly distributed with Italian PortMEDIA. 
This dataset was designed for portability purposes with the French MEDIA corpus, its dialogues being close-domain phone calls of spectacle booking. 
The corpus shares 19 semantic annotations with MEDIA, and integrates 17
others for the spectacle task, allowing us to evaluate both CER and CVER. 

Despite highly degrading French scores compared to the $11.5$\% WER $18.5$\% CER and $29.5$\% CVER obtained of MEDIA with SAMU-XLSR $_{IT \oplus FR}$, Table~\ref{tab:frpm} demonstrates significant score improvement when freezing or fine-tuning specialized speech encoders, with SAMU-XLSR $_{IT \oplus FR}$ leading to almost all best results in WER and CER, clearly surpassing SAMU-XLSR.
We can positively state that such specialization can enhance both semantics extraction and audio transcription for a close-domain task like French PortMEDIA. 

\begin{table}[ht]
    \begin{center}
        \vspace{-0.1cm}
        \caption{Inference results (\%) for domain portability from MEDIA to French PortMEDIA.}
        \vspace{-0.42cm}
        \resizebox{0.99\columnwidth}{!}{
        \begin{tabular}{ l  l | c | c | c |}
            \cline{3-5}   
            & & \textbf{WER} & \textbf{CER} & \textbf{CVER} \\
            \hline
            \multirow{4}{*}{Frozen} 
            & XLS-R & $47.5$ & $62.9$ & $73.9$ \\
            & SAMU-XLSR & $47.3$ & $61.5$ & $74.3$ \\
            & SAMU-XLSR $_{FR}$ & $47.5$ & $62.5$ & $75.3$ \\
            & SAMU-XLSR $_{IT \oplus FR}$ & \textbf{39.3} & \textbf{58.5} & \textbf{68.4} \\
            \hline
            \multirow{4}{*}{Fine-Tuned} 
            & XLS-R & $40.6$ & $58.2$ & $67.7$ \\
            & SAMU-XLSR & $38.6$ & $57.1$ & $66.1$ \\
            & SAMU-XLSR $_{FR}$ & $38.7$ & \textbf{56.0} & \textbf{65.7} \\
            & SAMU-XLSR $_{IT \oplus FR}$ & \textbf{32.8} & $56.1$ & $66.1$ \\
            \hline  
        \end{tabular}
        }
    \label{tab:frpm}
    \vspace{-0.8cm}
    \end{center}
\end{table}


\vspace{-0.11cm}
\section{Conclusions}
\label{sec:conclu}
\vspace{-0.05cm}

This paper investigates how to effectively enrich the frame-level speech representations of an SSL model for a complex SLU task, by proposing a specialized semantics injection with the recently introduced SAMU-XLSR approach. 
Each specialized SAMU-XLSR is then used as a speech encoder for a challenging semantic extraction task, with the low-resource Italian PortMEDIA benchmark.
After conducting their layer-wise capacity analysis, the paper experiments with both their cross-lingual and cross-domain portability. 
Results show how this enrichment of SAMU-XLSR can significantly improve semantic concept extraction, leading to the newly state-of-the-art $25.1$\% CER obtained on Italian PortMEDIA with language portability experiments. 
Analysis also demonstrate the usefulness of such specialisation for close cross-domain datasets.
One very interesting discovery also lies in the computation cost-efficiency of such specialized speech encoders, able to approximate our state-of-the-art results of a fully-fine-tuned SLU model, with only 17 of the 24 layers of a frozen SAMU-XLSR. 
Furthermore, the SSL model semantic enrichment done at sentence-level, rather than character or word level, opens promising perspectives to exploit imperfect transcriptions for the ASR and SLU domains.

\vspace{-0.11cm}
\section{Acknowledgements}
\label{sec:ack}
\vspace{-0.05cm}

This paper was inspired by insights gained from JSALT 2022 at JHU, gift-funded by Amazon, Microsoft and Google.
This work was performed using HPC resources from GENCI/IDRIS (grant 2022 AD011012565) and received funding from the EU H2020 SELMA (grant No 957017), ESPERANTO (grant No 101007666) and PSPC AIDA: 2019-PSPC-09 (funded by BPI-France) projects. 
We would like to especially thank our colleagues Sameer Khurana and Antoine Laurent for inventing SAMU-XSLR and inspiring us its use towards SLU.

\pagebreak
\bibliographystyle{IEEEtran}
\bibliography{mybib}

\end{document}